\pdfoutput=1

\documentclass[11pt]{article}

\usepackage{acl}

\usepackage{times}
\usepackage{latexsym}

\usepackage[T1]{fontenc}

\usepackage[utf8]{inputenc}

\usepackage{microtype}
\usepackage{graphicx}
\usepackage{covington}

\title{Do language models make human-like predictions about the coreferents of Italian anaphoric zero pronouns?}

\author{James A. Michaelov \\
  Department of Cognitive Science \\
  University of California, San Diego \\
  \texttt{j1michae@ucsd.edu} \And
  Benjamin K. Bergen \\
  Department of Cognitive Science \\
  University of California, San Diego \\
  \texttt{bkbergen@ucsd.edu} }

\begin{document}
\maketitle
\begin{abstract}
Some languages allow arguments to be omitted in certain contexts. Yet human language comprehenders reliably infer the intended referents of these \textit{zero} pronouns, in part because they construct expectations about which referents are more likely. We ask whether Neural Language Models also extract the same expectations. We test whether 12 contemporary language models display expectations that reflect human behavior when exposed to sentences with zero pronouns from five behavioral experiments conducted in Italian by \citet{carminati_2005_ProcessingReflexesFeature}. We find that three models---XGLM 2.9B, 4.5B, and 7.5B---capture the human behavior from all the experiments, with others successfully modeling some of the results. This result suggests that human expectations about coreference can be derived from exposure to language, and also indicates features of language models that allow them to better reflect human behavior.
\end{abstract}

\section{Introduction}
\label{sec:intro}
In Italian, like other \textit{pro-drop} (`pronoun-dropping') languages, verbal arguments that would usually be expressed by pronouns in languages such as English can be omitted under certain circumstances. For example, consider the sentence in (\ref{ex:carminati_ex}) from \citet{carminati_2005_ProcessingReflexesFeature}.

\begin{example}\label{ex:carminati_ex}
\textit{Quando Maria ha chiamato Mario, era contenta}.\\`When Maria called Mario, [she] was happy.'
\end{example}

In this sentence, the referent of the `dropped' pronoun---generally referred to as a \textit{zero} or \textit{null} pronoun---can be inferred from the fact that the adjective \textit{contenta} is feminine; thus, Maria is the most likely subject of the second clause. Resolving the referents of anaphoric zero pronouns like this is a long-standing, important, and active area of research in natural language understanding (\citealp{jurafsky_2021_SpeechLanguageProcessing}; for examples, see \citealp{zhao_2007_IdentificationResolutionChinese,taira_2008_JapanesePredicateArgument,imamura_2009_DiscriminativeApproachPredicateArgument,watanabe_2010_StructuredModelJoint,kong_2010_TreeKernelBasedUnified,poesio_2010_CreatingCoreferenceResolution,chen_2013_ChineseZeroPronoun,yoshino_2013_PredicateArgumentStructure,iida_2016_IntraSententialSubjectZero,aloraini_2020_CrosslingualZeroPronoun,song_2020_ZPR2JointZeroa,ueda_2020_BERTbasedCohesionAnalysis,konno_2020_EmpiricalStudyContextual,konno_2021_PseudoZeroPronoun,yang_2020_TransformerGCRFRecoveringChinesea,kim_2021_ZeroanaphoraResolutionKorean,umakoshi_2021_JapaneseZeroAnaphora,chen_2021_TacklingZeroPronoun,yang_2022_CorefDPRJointModel}).

It has been argued that aiming for human-like-ness in natural language processing systems is vital if we want our natural language understanding systems to behave not only as humans do, but also as human users expect them to \citep[see, e.g.,][]{keller_2010_CognitivelyPlausibleModels,ettinger_2020_WhatBERTNot,eisape_2020_ClozeDistillationImproving}. This is particularly true for zero anaphora resolution, and pronoun resolution more generally. As an illustration of the prominence of reference resolution, one pronoun resolution task, the Winograd Schema Challenge (\citealp{levesque_2012_WinogradSchemaChallenge}; based on work by \citealp{winograd_1972_UnderstandingNaturalLanguage}), has been referred to as `an alternative to the Turing Test' \citep[p.~552]{levesque_2012_WinogradSchemaChallenge}.

So, how do humans resolve coreference? The evidence suggests that we use a range of cues---for example, agreement information as in (\ref{ex:carminati_ex}), but also factors such as world knowledge and common sense \citep{winograd_1972_UnderstandingNaturalLanguage,hobbs_1979_CoherenceCoreference,kehler_2007_CoherenceCoreferenceRevisited,kehler_2013_ProbabilisticReconciliationCoherencedriven,sakaguchi_2019_WinoGrandeAdversarialWinograd}. In addition, pronoun resolution is shaped by our expectations about the next entity that is likely to be mentioned and what argument it should take \citep{kehler_2007_CoherenceCoreferenceRevisited,kehler_2013_ProbabilisticReconciliationCoherencedriven,nieuwland_2006_IndividualDifferencesContextual}. For example, crosslinguistic work has found a bias towards expecting a subject pronoun to refer to an antecedent subject \citep[for discussion, see][]{carminati_2005_ProcessingReflexesFeature}. This has been demonstrated experimentally with sentences such as those in (\ref{ex:stephenson}). 

\begin{example}\label{ex:stephenson}
\textit{John seized the comic from Bill. He\_\_\_\_}
\end{example}

When presenting experimental participants with sentences such as  (\ref{ex:stephenson}) where the content following the pronoun has been removed, \citet{stevenson_1994_ThematicRolesFocus} found that the vast majority of people expect \textit{he} to refer to \textit{John} rather than \textit{Bill}. The effect of expectations such as these are so powerful that we may often not even realize that a sentence is grammatically ambiguous at all in most situations \citep{nieuwland_2006_IndividualDifferencesContextual}.

The same principles apply in zero anaphora resolution. \citet{carminati_2005_ProcessingReflexesFeature}, for example, tests human expectations by investigating how long it takes for experimental participants to read stimuli with certain linguistic features, based on the well-established finding that contextually expected words are read faster than unexpected words, demonstrating an increased processing difficulty when these expectations are violated \citep[see, e.g.][]{forster_1981_PrimingEffectsSentence,levy_2008_ExpectationbasedSyntacticComprehension,luke_2016_LimitsLexicalPrediction,brothers_2021_WordPredictabilityEffects}. \citet{carminati_2005_ProcessingReflexesFeature} finds that the main clauses of sentences such as (\ref{ex:carminati_ex})---that is, the part of the sentence containing the zero subject pronoun, i.e., \textit{era contenta} (`[she] was happy')---are read faster when the zero pronoun co-refers with a subject antecedent, as in (\ref{ex:carminati_ex}), than when it  co-refers with the antecedent object, as in (\ref{ex:carminati_ex2}). 

\begin{example}\label{ex:carminati_ex2}
\textit{Quando Maria ha chiamato Mario, era contento}.\\`When Maria called Mario, [he] was happy.'
\end{example}

The question, then, if we want human-like zero anaphora resolution, is how to test whether a given zero anaphora resolution system is able to reflect these human expectations. In the present study, we propose a method to do just that.

The vast majority of recent pronoun resolution systems base their approach around using the representations learned by contemporary transformer language models---for example, in the zero pronoun anaphora resolution literature alone, researchers have used pretrained transformers such as monolingual \citep{song_2020_ZPR2JointZero,ueda_2020_BERTbasedCohesionAnalysis,konno_2020_EmpiricalStudyContextual,konno_2021_PseudoZeroPronoun,kim_2021_ZeroanaphoraResolutionKorean,chen_2021_TacklingZeroPronoun,umakoshi_2021_JapaneseZeroAnaphora} and multilingual \citep{aloraini_2020_CrosslingualZeroPronoun,kim_2021_ZeroanaphoraResolutionKorean} BERT models \citep{devlin_2019_BERTPretrainingDeep}, as well as XLM-R (\citealp{conneau_2020_UnsupervisedCrosslingualRepresentation}; for an example see \citealp{yang_2022_CorefDPRJointModel}). 

For these systems, there is a clear way to test for human-like-ness. We can directly probe the extent to which the representations learned by the language models take into account the factors that lead to coreference expectations in humans by testing how similar the predictions of language models are to those of human comprehenders---if they exhibit the same pattern of predictive behavior as humans in a given context, this demonstrates that they are sensitive to the same factors as humans in this context. We do this by comparing the reading times reported by \citet{carminati_2005_ProcessingReflexesFeature} to the surprisals of 12 contemporary transformer language models \citep{devlin_2019_BERTPretrainingDeep,conneau_2019_CrosslingualLanguageModel,demattei_2020_GePpeTtoCarvesItalian,schweter_2020_ItalianBERTELECTRA,conneau_2020_UnsupervisedCrosslingualRepresentation,devries_2021_GoodNewHow,lin_2021_FewshotLearningMultilinguala} for the same stimuli.

A key question is whether it is possible for language models to learn any of these human-like expectations at all, given that they can only rely on the statistics of language. For this reason, the results of the present study should be of interest both from a natural language understanding perspective, as discussed above, and also from a psycholinguistics perspective.

From the natural language understanding perspective, the present study presents an approach for `pre-evaluating' a language model's suitability as a basis for a zero anaphora resolution system. Specifically, if a language model can model a specific effect in human language processing---that is, if an experimental manipulation that elicits a significant difference in reading time also results in a significant difference in that language model's surprisal in the same direction---this demonstrates that it is able to take into account the relevant factors that underlie human comprehender expectation. For example, if a language model can successfully model the subject antecedent preference, this suggests that it has learned that all else being equal, subject antecedents are more likely to be the coreferents of zero subject pronouns, and thus, crucially, that this pattern is in some way represented in the contextual embeddings that can be used as the representations underlying a zero anaphora resolution system.

From the psycholinguistics perspective, this study explores the extent to which it is possible that specific patterns in zero anaphora coreference expectations can be learned on the basis of the statistics of language alone. There is substantial work demonstrating that some expectations are highly correlated with language statistics, and thus may be at least partly derived from them \citep{levy_2008_ExpectationbasedSyntacticComprehension,monsalve_2012_LexicalSurprisalGeneral,smith_2013_EffectWordPredictability,frank_2015_ERPResponseAmount,michaelov_2020_HowWellDoes,szewczyk_2022_ContextbasedFacilitationSemantic}. However, other work has suggested that coreference expectations are instead (or in addition) at least partly based on semantic knowledge, world experience, and conceptual salience \citep{hobbs_1979_CoherenceCoreference,harley_2002_PersonNumberPronouns,carminati_2005_ProcessingReflexesFeature,kehler_2007_CoherenceCoreferenceRevisited,kehler_2013_ProbabilisticReconciliationCoherencedriven}. Nonetheless, since the predictions of language models are derived from language statistics alone, if even one language model can successfully model a given effect (after adjusting for multiple comparisons), this provides in-principle evidence that the effect can be successfully learned using distributional information alone.

\section{General Method}
\label{sec:methods}
The experiments reported by \citet{carminati_2005_ProcessingReflexesFeature} were self-paced reading experiments. Participants were native speakers of Italian asked to read Italian sentences on a computer. Stimuli were similar to those discussed in the previous section, with a subordinate clause (e.g. \textit{Quando Maria ha chiamato Mario}; `When Maria called Mario') first presented, followed by the main clause (e.g. either \textit{era contenta} `[she] was happy' or \textit{era contento} `[he] was happy'). The time taken by participants to read the main clause---which includes the word that disambiguates the null subject pronoun---was recorded.

To measure the language model's expectations, we used surprisal (negative log-probability) based on a large body of evidence that language model surprisal generally correlates well with reading time \citep[see, e.g.][]{levy_2008_ExpectationbasedSyntacticComprehension,monsalve_2012_LexicalSurprisalGeneral,smith_2013_EffectWordPredictability,goodkind_2018_PredictivePowerWord} and other metrics of processing difficulty that are thought to correlate with human expectations such as the neural N400 response \citep{frank_2015_ERPResponseAmount,aurnhammer_2019_EvaluatingInformationtheoreticMeasures,michaelov_2020_HowWellDoes,merkx_2021_HumanSentenceProcessing}. 

To model each effect, we compared whether specific linguistic features of the stimuli that elicited a significant difference in human reading times also led to a significant difference in language model surprisal. For example, we investigate whether, like reading time, surprisal is significantly lower when the referent of a zero subject pronoun is a subject antecedent compared to an object antecedent, among other patterns in reading time reported by \citet{carminati_2005_ProcessingReflexesFeature}. The language models were all presented with the same stimuli as the human participants, which are provided by \citet{carminati_2005_ProcessingReflexesFeature} in an appendix to the original paper. 

To match reading time, surprisal was calculated over the whole of the main clause in each stimulus item. This was done by calculating the sum of the surprisals of the main clauses' constituent words, which is equivalent to taking the negative logarithm of the product of their probabilities.

We ran the stimuli through 12 transformer language models---5 monolingual and 7 multilingual. Two of the monolingual models were autoregressive transformer networks: GePpeTto \citep{demattei_2020_GePpeTtoCarvesItalian} and the small English GPT-2 retrained on Italian \citep{devries_2021_GoodNewHow}. The three remaining monolingual models were masked language models: UmBERTo \citep{parisi_2021_UmBERToCommoncrawlCased} trained on the Italian subcorpus of OSCAR, and the Base and XXL versions of the Italian BERT models \citep{schweter_2020_ItalianBERTELECTRA}. The multilingual models also included autoregressive and masked language models. The autoregressive models were three different sizes of XGLM \citep{lin_2021_FewshotLearningMultilinguala}: the 2.9B, 4.5B, and 7.5B parameter models. The masked language models were XLM-100 \citep{conneau_2019_CrosslingualLanguageModel}, and the Base and Large versions of XLM-R \citep{conneau_2020_UnsupervisedCrosslingualRepresentation}. 

The aim in using this range of models was to test whether there are any model types or characteristics made them better suited to capturing human behavior---for instance, whether the models were autoregressive or masked, or monolingual or multilingual. Previous systems designed to resolve zero pronoun anaphora of the kind described here appear to be predominantly based on masked language models; however, autoregressive models such as GPT-2 have been successfully used in similar systems \citep{maqbool_2022_ZerolabelAnaphoraResolution}. 

We are also interested in whether monolingual or multilingual models are better suited to the task of zero pronoun anaphora resolution---while cross-lingual transfer may help with some phenomena \citep{guarasci_2022_BERTSyntacticTransfer}, there is also evidence that it can cause harm to model performance in others \citep{wang_2020_NegativeInterferenceMultilingual}. There is currently mixed evidence with respect to zero pronoun anaphora resolution---\citet{kim_2021_ZeroanaphoraResolutionKorean}, for example, find that a monolingual Korean BERT-based model performs better than the standard multilingual BERT model; while \citet{yang_2022_CorefDPRJointModel} finds that their model, based on XLM-R, is better than a model based on a Chinese-only BERT \citep{song_2020_ZPR2JointZero}. We include both multilingual BERT and XLM-R in our analyses, in addition to the Base Italian BERT model, which has previously been evaluated in terms of its capacity to learn non-anaphoric null subject and agreement phenomena in Italian \citep{guarasci_2021_AssessingBERTAbility}.

To test whether each model successfully modeled each effect, we constructed linear mixed-effects models predicting model surprisal with experimental manipulation as a main effect and a random intercept of sentence frame, where sentence frame refers to a set of stimuli that differ only by experimental condition (e.g., the previously discussed \textit{Quando Maria ha chiamato Mario, era contenta} and \textit{Quando Maria ha chiamato Mario, era contento} are two stimuli with the same sentence frame). 

For three of the five analyses---the two analyses in Section \ref{ssec:subj_results} where the coreferent is distinguished by gender, and the analysis in Section \ref{ssec:name_pronoun}---we tested whether the relevant experimental manipulation was a significant predictor of language model surprisal by constructing a null regression with only the random intercept of sentence frame and running a likelihood ratio test investigating whether adding the experimental manipulation improved model fit. 

The remaining two analyses correspond to two different tests utilized by \citet{carminati_2005_ProcessingReflexesFeature} to analyze the results of a single experiment (Experiment 4 of the original paper). Crucially, \citet{carminati_2005_ProcessingReflexesFeature} tests whether there is an interaction between coreferent argument (whether it is the antecedent subject or object) and coreferent person (whether the coreferent is in the first or second person or in the third person), but also whether there is a main effect of each of these. To test whether there is an interaction (in Section \ref{ssec:results_person}), we construct a linear mixed-effects model with and without the interaction, and run a likelihood ratio test comparing the two. In addition to the interaction,  \citet{carminati_2005_ProcessingReflexesFeature} finds a main effect of coreferent argument but not of person. Thus, we also test for the main effect of coreferent argument, which we report in Section \ref{ssec:subj_results}. Because we want to investigate whether the main effect of coreferent argument explains a significant amount of the variance in surprisal while also accounting for the effect of a possible interaction, instead of using a likelihood ratio test, we opt for a Type III ANOVA with Satterthwaite's method for estimating degrees of freedom \citep{kuznetsova_2017_LmerTestPackageTests}. 

\begin{figure*}[!h]
    \centering
    \includegraphics[width=0.93\textwidth]{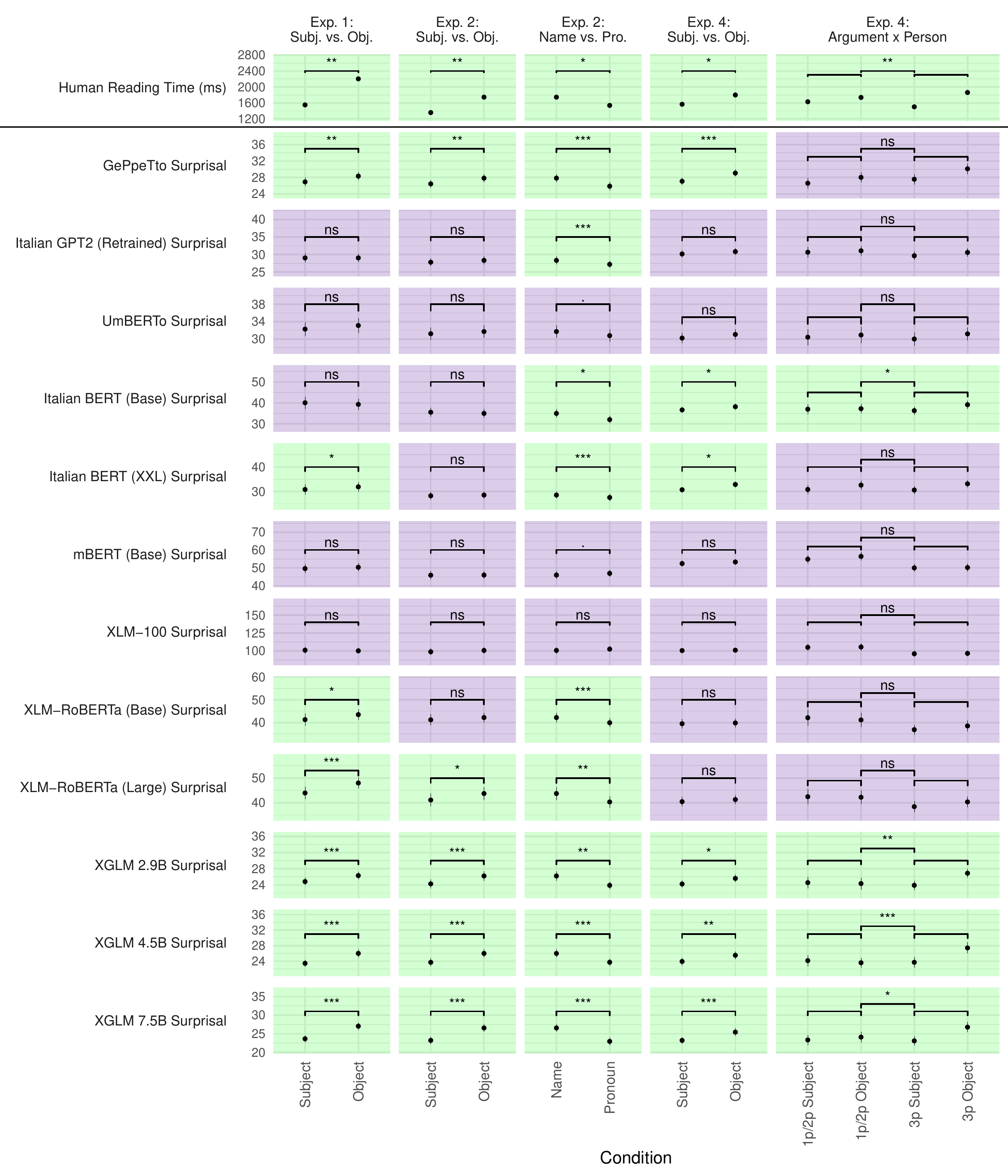}
    \caption{Mean reading time and surprisal of each model elicited by main clauses for each experimental condition in each experiment. All significant differences are shown: following convention, `***' indicates $p<0.001$, `**' indicates $p<0.01$, `*' indicates $p<0.5$, `.' indicates marginal significance where $p<0.1$, and `ns' indicates $p\geq0.1$. For easier comparison across models and experiments, comparisons with statistically significant results are colored green; non-significant results are colored purple. Note that the relevant $p$-values have been corrected for multiple comparisons using the method of \citet{benjamini_1995_ControllingFalseDiscovery}; for test statistics and degrees of freedom, see \autoref{sec:appendix}. Details of the statistical tests for reading time are provided by \citet{carminati_2005_ProcessingReflexesFeature}. For language model surprisal, error bars indicate standard error; no metric of error is provided by \citet{carminati_2005_ProcessingReflexesFeature}.}
    \label{fig:all}
\end{figure*}

The details of the results of the statistical analyses that were run by \citet{carminati_2005_ProcessingReflexesFeature} are provided in the original paper. The full results of the statistical analyses that we ran are provided in \autoref{sec:appendix}. The results of both sets of statistical analyses are summarized in \autoref{fig:all}.

All language models were run in \textit{Python} \citep{vanrossum_2009_PythonReferenceManual}, using the \textit{PyTorch} \citep{paszke_2019_PyTorchImperativeStyle} implementation of each model, as provided by the \textit{transformers} package \citep{wolf_2020_TransformersStateoftheArtNatural}. Statistical analysis and data manipulation were carried out in \textit{R} \citep{rcoreteam_2020_LanguageEnvironmentStatistical} using \textit{Rstudio} \citep{rstudioteam_2020_RStudioIntegratedDevelopment} and the \textit{tidyverse} \citep{wickham_2019_WelcomeTidyverse}, \textit{lme4} \citep{bates_2015_FittingLinearMixedeffects}, \textit{lmerTest} \citep{kuznetsova_2017_LmerTestPackageTests}, \textit{ggsignif} \citep{ahlmann-eltze_2021_GgsignifPackageDisplaying}, and \textit{cowplot} \citep{wilke_2020_CowplotStreamlinedPlot} packages. The stimuli, code used to run the models, and code used to run the statistical analyses are provided on Github\footnote{\url{https://github.com/jmichaelov/italian-zero-anaphora-prediction}}.  Note that all $p$-values reported in this analysis have been corrected for multiple comparisons \citep{benjamini_1995_ControllingFalseDiscovery,rcoreteam_2020_LanguageEnvironmentStatistical}.

\section{Manipulation-level results and discussion}
In this section, we compare the performance of the 12 language models tested with human behavior on five of the experimental manipulations carried out by \citet{carminati_2005_ProcessingReflexesFeature}. Note that two additional studies from that paper focus on a different question---the effects of distractor referents on processing time. Although at least one model was able to capture each of these human results, they are not included here because they address a different set of phenomena.

\subsection{Subject vs. object antecedent referent}
\label{ssec:subj_results}
\citet{carminati_2005_ProcessingReflexesFeature} investigates the subject antecedent preference discussed in Section \ref{sec:intro} in three experiments. In Experiments 1 and 2 of the original paper, both antecedents are names associated with different genders, as illustrated by the example from Experiment 1 shown in (\ref{ex:carminati_exp1}).

\begin{subexamples}
\label{ex:carminati_exp1}
\item \textit{Quando Lucia ha telefonato a Marco, era appena tornata da Londra.}\\`When Lucia has telephoned Marco, [she] had just come back from London.'
\item \textit{Quando Lucia ha telefonato a Marco, era appena tornato da Londra.}\\`When Lucia has telephoned Marco, [he] had just come back from London.'
\end{subexamples}

Because \textit{tornato/tornata} (`come back') agrees with the gender of the zero subject pronoun, its referent can be resolved to be the subject antecedent (\textit{Lucia}) in (4a) and the object antecedent (\textit{Marco}) in (4b). \citet{carminati_2005_ProcessingReflexesFeature} found, as expected, that main clauses where the zero subject pronoun co-referred with the subject antecedent (like (4a)) were read faster than those where they had an object antecedent coreferent (like (4b)), suggesting an expectation for a subject antecedent coreferent. 

In Experiment 4 of the original paper, grammatical person was manipulated rather than grammatical gender, as illustrated by the example in (\ref{ex:carminati_exp4_short}).

\begin{subexamples}
\label{ex:carminati_exp4_short}
\item \textit{Quando ho litigato con Maria, ero molto prepotente.}\\`When [I] quarrelled with Maria, [I] was very pushy.'
\item \textit{Quando ho litigato con Maria, era molto prepotente.}\\`When [I] quarrelled with Maria, [she] was very pushy.'
\end{subexamples}

Similarly, because \textit{ero/era} (`was') either agrees with the first person or third person, the zero subject pronoun can be resolved as co-referring with the speaker (in (5a)) or with Maria (in (5b)). As in the aforementioned other experiments, \citet{carminati_2005_ProcessingReflexesFeature} finds that speakers read sentences like (5a) faster than sentences like (5b), again demonstrating a preference for subject antecedent coreferents over object antecedent coreferents.

Looking at the results of the models, we can see that only GePpeTto and the XGLMs successfully model this effect in all three experiments. This appears to suggest that autoregressive models may be better at learning that the subject antecedent is the more likely referent; however, it should be noted that in each of the individual studies, at least one masked language model also successfully modeled the effect. Nonetheless, the robustness of similarity between these autoregressive models' predictions and human expectations may be partly explained by the evidence suggesting that autoregressive models are more sensitive to word order than masked language models, to the extent that they are able to encode positional information even without explicit positional encodings (\citealp{haviv_2022_TransformerLanguageModels}); conversely, masked language models appear to be relatively insensitive to word order \citep{sinha_2021_MaskedLanguageModeling,gupta_2021_BERTFamilyEat}. Given that the dominant pattern in Italian is Subject-Verb-Object \citep[see][]{guarasci_2022_BERTSyntacticTransfer} and the subject was always first in the subordinate clause, it is therefore unsurprising that autoregressive models would be better able to predict that the first entity mentioned (the subject) is more likely as the subject of the zero pronoun than the second entity mentioned (the object).

\subsection{Name vs. pronoun antecedent referent}
\label{ssec:name_pronoun}
In addition to investigating the differences in how humans process zero anaphora in sentences with subject and object antecendent coreferents, \citet{carminati_2005_ProcessingReflexesFeature} also investigated how the form in which antecedents are presented impacts processing. As a further part of Experiment 2 of the original paper, \citet{carminati_2005_ProcessingReflexesFeature} investigates how processing is impacted when the object coreferent is presented as a name or a pronoun, an example of which is provided in (\ref{ex:carminati_exp2.2}).

\begin{subexamples}
\label{ex:carminati_exp2.2}
\item \textit{Quando Maria cerca Roberto, diventa ansioso.}\\`When Maria looks for Roberto, [he] becomes anxious.'
\item \textit{Quando Maria lo cerca, diventa ansioso.}\\`When Maria looks for him, [he] becomes anxious.'
\end{subexamples}

In both sentences, it is the object antecedent that is the referent of the zero pronoun in the main clause, violating the subject antecedent preference. \citet{carminati_2005_ProcessingReflexesFeature} finds that main clauses with zero pronouns referring to antecedent objects are easier to process (read faster) when this antecedent object is a pronoun.

The results for the language models, shown in \autoref{fig:all}, suggest that this is a relatively easy pattern for language models to learn---9 of the 12 models show a significant effect and the remaining 3 show a marginal effect in the correct direction. Thus it is clear that this general rule---that an entity referred to by an antecedent pronoun is more likely to be the referent of a zero pronoun---is possible to learn based on the statistics of language. The fact that this effect relies on the form of the antecedents rather than word order could explain why there is no difference between autoregressive and masked language models in this case.

\subsection{Antecedent argument by grammatical person interaction}
\label{ssec:results_person}
In addition to investigating the subject antecedent effect, in Experiment 4 of the original paper, \citet{carminati_2005_ProcessingReflexesFeature} investigates how this effect interacts with the the grammatical person of antecedents (i.e., first, second, or third-person). In general, previous work suggests that first and second-person antecedents are more likely to be referents of reduced or zero pronouns \citep{ariel_1991_FunctionAccessibilityTheory,siewierska_1999_ReducedPronominalsArgument,siewierska_2003_ReducedPronominalsArgument,carminati_2005_ProcessingReflexesFeature}, but as has been discussed, subject antecedents are also more likely to be their referents. Thus, \citet{carminati_2005_ProcessingReflexesFeature} compares the effect of the person of the coreferent antecedent when it is in both subject and object position, as exemplified in (\ref{ex:carminati_exp4}).

\begin{subexamples}
\label{ex:carminati_exp4}
\item \textit{Quando ho/hai litigato con Maria, ero/eri molto prepotente.}\\`When [I/you] quarrelled with Maria, [I/you] was/were very pushy.'
\item \textit{Quando Maria ha litigato con me/te, ero/eri molto prepotente.}\\`When Maria quarrelled with me/you, [I/you] was/were very pushy.'
\item \textit{Quando Maria ha litigato con me/te, era molto prepotente.}\\`When Maria quarrelled with me/you, [she] was very pushy.'
\item \textit{Quando ho/hai litigato con Maria, era molto prepotente.}\\`When [I/you] quarrelled with Maria, [she] was very pushy.'
\end{subexamples}

While \citet{carminati_2005_ProcessingReflexesFeature} does not find a main effect of grammatical person, the results show an interaction between person and antecedent referent argument status (i.e. whether it is a subject or object). Specifically, the difference in reading time between subject and object antecedent referents is reduced when the antecedent coreferent is in the first or second person. In other words, the subject antecedent effect is weaker with first and second person coreferents. This, \citet{carminati_2005_ProcessingReflexesFeature} argues, shows that the bias towards a first or second-person coreferent modulates the bias against an object coreferent---in other words, humans still expect a first or second-person coreferent even if it is an object antecedent. 

Four of the models---Italian BERT Base and the XGLMs---manage to model this interaction. While this suggests the the effect---which is complicated as it relies on correctly weighting the effects of argument status and person---is difficult to learn based on the statistics of language, it nevertheless demonstrates that it is indeed possible.

\section{General Discussion}

\subsection{Implications for human language processing}
We can now return to the two questions that motivated this work. First, we look at whether the reading time effects in humans can be explained on the basis of the statistics of language.

As seen in \autoref{fig:all}, each experimental result was successfully modeled by at least four language models, after correcting for multiple comparisons. This shows that it is possible to learn cues based on the statistics of language that result in human-like expectations about the referents of zero subject pronouns in Italian. The fact that the XGLM transformers were consistently able to model all the effects demonstrates that the patterns underlying the results of the experiments can all be learned by the same system---and therefore, in principle, it should also be possible for a neurocognitive system implementing lexical prediction in humans \citep[for accounts of such a system and what it might learn, see, e.g.,][]{kutas_2011_LookWhatLies,lewis_2015_PredictiveCodingFramework,lupyan_2015_WordsWorldPredictive,frank_2015_ERPResponseAmount,bornkessel-schlesewsky_2019_NeurobiologicallyPlausibleModel,aurnhammer_2019_EvaluatingInformationtheoreticMeasures,michaelov_2020_HowWellDoes,kuperberg_2020_TaleTwoPositivities,merkx_2021_HumanSentenceProcessing,brothers_2021_WordPredictabilityEffects}. Thus, the present study provides evidence that the expectations that humans form about possible referents in anaphora may be derived from language statistics, at least in part.

\subsection{Implications for work on language models}

\begin{table}[h]
\centering
\begin{tabular}{lr}
\hline
\textbf{Model} & \textbf{Experiments modeled}\\
\hline
GePpeTto & 4/5\\
It. GPT2 (Retrained)& 1/5\\
UmBERTo & 0/5\\
It. BERT (Base)& 3/5 \\
It. BERT (XXL)& 3/5\\
mBERT& 0/5\\
XLM-100& 0/5 \\
XLM-R (Base)& 2/5\\
XLM-R (Large)& 3/5 \\
XGLM 2.9B& 5/5 \\
XGLM 4.5B& 5/5 \\
XGLM 7.5B& 5/5 \\
\hline
\end{tabular}
\caption{Number of experiments successfully modeled by each language model.}
\label{tab:disc}
\end{table}

The number of experiments successfully modeled by each language model is shown in \autoref{tab:disc}, revealing that the XGLM models performs best overall, successfully modeling the results of all 5 experiments investigated. After the XGLMs, GePpeTto models the most experiments (4/5), followed by XLM-R Large and the Italian BERTs (3/5). The remaining transformers only successfully model 2 or fewer of the experiments.

At this level of analysis, some patterns begin to emerge. First, the best models are the XGLM transformers and GePpeTto. This suggests that autoregressive models may in fact be best able to model the effects. As discussed in Section \ref{ssec:subj_results}, this may be due to their comparatively high sensitivity to word order. One issue that confounds this interpretation is that the XGLM models are also larger and trained more data on than the other models. However, the fact that GePpeTto was trained on 13GB of text, while the other monolingual models (which were all masked language models) were trained on the same amount or more data and performed worse, suggests that, at the very least, monolingual autoregressive models may more efficiently learn biases in zero anaphora processing than monolingual masked language models. Whether or not autoregressive models continue to out-perform masked language models as the training set increases in size is a question for further research. Overall, then, we see that in our sample of models, autoregressive monolingual and multilingual models are more human-like in their expectations of zero subject pronoun referents than their masked language model counterparts.

Another question that we can address with the present results is that of the effect of multilinguality on the human-likeness of the models' expectations. First, while GePpeTto and Retrained Italian GPT-2 are trained on the same Italian corpus, the former greatly out-performs the latter. This suggests that training a model on one language and then re-training it on another does not necessarily improve the representations that a model learns---in fact, in this case, it interferes with the model's ability to make predictions in a human-like fashion. On the other hand, XLM-R Large is trained on data from 100 languages successfully models human processing at least as well as any monolingual model but GePpeTto---including Italian BERT XXL, which is trained on 80GB of Italian text compared to XLM-R's 30GB. Thus, it may be the case that with more training data, and with a larger number of languages (including more closely-related languages---XLM-R is also trained on other Romance languages), there is some cross-linguistic transfer that can aid in predicting the referent of a null subject pronoun in a human-like manner \citep[see][for a recent similar finding]{guarasci_2022_BERTSyntacticTransfer}. Finally, the XGLMs---autoregressive multilingual models---are the best performing models overall. Thus, the results of this study seem to suggest that with enough overall data, and when multilingual language models are trained on more languages, cross-linguistic transfer can improve their human-likeness in terms of their predictions. A question for future work is to investigate under what circumstances multilinguality hurts or harms the human-likeness of language model predictions---for example, based on how related the languages the model is trained on are to each other, or how widespread the phenomenon under investigation is. For example, the subject antecedent preference is also present in English with overt pronoun anaphora \citep{smyth_1994_GrammaticalDeterminantsAmbiguous,chambers_1998_StructuralParallelismDiscourse,kehler_2007_CoherenceCoreferenceRevisited,kehler_2013_ProbabilisticReconciliationCoherencedriven}.

Finally, as discussed in Section \ref{sec:methods}, \citet{yang_2022_CorefDPRJointModel} show that a zero pronoun anaphora resolution system based on XLM-R performs better than one based on multilingual BERT \citep{song_2020_ZPR2JointZero}. Concurrently, in the present study, we see that either XLM-R model is better able to model zero anaphora processing effects than multilingual BERT. While there are other factors at play, this result is consistent with our prediction that better modeling of human expectations may lead to better performance when using the models' representations for zero pronoun anaphora resolution, based on the idea that the representations learned by the model better allow it to make human-like predictions, and thus are more useful for systems aiming to resolve zero anaphora in a human-like way. In the present study, XGLM models perform better than the other models, and thus, based on this, we suggest that XGLM transformers may be better models upon which to base future zero pronoun anaphora resolution system than other current publicly available pretrained models.

\section{Conclusion}
We present the first study investigating whether language models make the same predictions as humans when processing zero pronoun anaphora. For each the 5 effects we investigate, we find that there are at least four models that successfully do so; and three models, XGLM 2.9B, 4.5B, and 7.5B, successfully do so in all 5. This suggests that human processing of zero pronoun anaphora may at least partly rely on our statistical knowledge of language. Furthermore, this approach provides a useful way to investigate how human-like the referent predictions of language models are, which is vital if we are to use their representations for zero anaphora resolution systems.

\section*{Acknowledgements}
We would like to thank Maria Nella Carminati for making her stimuli available, and the anonymous reviewers for their helpful comments. We would also like to thank the other members of the Language and Cognition Lab at UCSD for their valuable discussion and the University of California, and the San Diego Social Sciences Computing Facility Team for technical assistance. This work was partially supported by a 2021-2022 Center for Academic Research and Training in Anthropogeny Annette Merle-Smith Fellowship awarded to James A. Michaelov, and the RTX A5000 used for this research was donated by the NVIDIA Corporation.

\bibliography{library}
\bibliographystyle{acl_natbib}

\appendix
\section{Full results of statistical analyses}
\label{sec:appendix}

\subsection*{Experiment 1: Subject vs. object antecedent referent}

\begin{table}[!h]
\centering
\begin{tabular}{lrr}
\hline
\textbf{Model}       & \textbf{Chisq(df=1)} & \textbf{Corrected \textit{p}}                \\ \hline
\textbf{GePeTto}     & \textbf{12.3}        & \textbf{0.002}            \\
GPT-2 Italian        & \textless{}0.1       & 0.993                     \\
UmBERTo              & 3.5                  & 0.109                     \\
It BERT Base         & 0.4                  & 0.614                     \\
\textbf{It BERT XXL} & \textbf{6.3}         & \textbf{0.025}            \\
mBERT                & 1.4                  & 0.376                     \\
XLM-100              & 0.2                  & 0.758                      \\
\textbf{XLM-R Base}  & \textbf{5.3}         & \textbf{0.041}            \\
\textbf{XLM-R Large} & \textbf{19.2}        & \textbf{\textless{}0.001} \\
\textbf{XGLM 2.9B}   & \textbf{13.9}         & \textbf{\textless{}0.001}            \\
\textbf{XGLM 4.5B}   & \textbf{19.1}          & \textbf{\textless{}0.001}            \\
\textbf{XGLM 7.5B}   & \textbf{32.9}        & \textbf{\textless{}0.001} \\ \hline
\end{tabular}
\caption{Results of the likelihood ratio tests in Experiment 1. Models for which there is a significant effect of the manipulation tested are shown in bold.}
\label{tab:exp1}
\end{table}

\subsection*{Experiment 2: Subject vs. object antecedent referent}

\begin{table}[!h]
\centering
\begin{tabular}{lrr}
\hline
\textbf{Model}       & \textbf{Chisq(df=1)} & \textbf{Corrected \textit{p}}                \\ \hline
\textbf{GePeTto}     & \textbf{8.8}         & \textbf{0.008}            \\
GPT-2 Italian        & 0.9                  & 0.482                     \\
UmBERTo              & 0.7                  & 0.514                     \\
It BERT Base         & 0.9                  & 0.482                     \\
It BERT XXL          & 0.6                  & 0.570                      \\
mBERT                & \textless{}0.1       & 0.956                     \\
XLM-100              & 0.9                  & 0.482                     \\
XLM-R Base           & 1                    & 0.468                     \\
\textbf{XLM-R Large} & \textbf{7.5}         & \textbf{0.015}            \\
\textbf{XGLM 2.9B}   & \textbf{17.5}         & \textbf{\textless{}0.001}            \\
\textbf{XGLM 4.5B}   & \textbf{23.6}         & \textbf{\textless{}0.001}            \\
\textbf{XGLM 7.5B}   & \textbf{26.5}        & \textbf{\textless{}0.001} \\ \hline
\end{tabular}
\caption{Results of the likelihood ratio tests for all models in Experiment 2.1. Models for which there is a significant effect of the manipulation tested are shown in bold.}
\label{tab:exp2_ab}
\end{table}

\newpage

\subsection*{Experiment 2: Name vs. pronoun object antecedent referent}

\begin{table}[!h]
\centering
\begin{tabular}{lrr}
\hline
\textbf{Model}         & \textbf{Chisq(df=1)} & \textbf{Corrected \textit{p}}                \\ \hline
\textbf{GePeTto}       & \textbf{29.8}        & \textbf{\textless{}0.001} \\
\textbf{GPT-2 Italian} & \textbf{17.3}        & \textbf{\textless{}0.001} \\
UmBERTo                & 4.9                  & 0.050                     \\
\textbf{It BERT Base}  & \textbf{7.6}         & \textbf{0.015}            \\
\textbf{It BERT XXL}   & \textbf{17.5}        & \textbf{\textless{}0.001} \\
mBERT                  & 4.3                  & 0.068                     \\
XLM-100                & 2.7                  & 0.168                    \\
\textbf{XLM-R Base}    & \textbf{14.7}        & \textbf{\textless{}0.001} \\
\textbf{XLM-R Large}   & \textbf{11.7}        & \textbf{0.002}            \\
\textbf{XGLM 2.9B}     & \textbf{12.4}        & \textbf{0.002}  \\
\textbf{XGLM 4.5B}     & \textbf{16.4}        & \textbf{\textless{}0.001}            \\
\textbf{XGLM 7.5B}     & \textbf{26.6}        & \textbf{\textless{}0.001} \\ \hline
\end{tabular}
\caption{Results of the likelihood ratio tests for all models. Models for which there is a significant effect of the manipulation tested are shown in bold.}
\label{tab:exp2_bc}
\end{table}

\subsection*{Experiment 4: Subject vs. object antecedent referent}

\begin{table}[!h]
\centering
\begin{tabular}{lrr}
\hline
\textbf{Model}        & \textbf{F(1,60)} & \textbf{Corrected \textit{p}}                \\ \hline
\textbf{GePeTto}      & \textbf{34.6}    & \textbf{\textless{}0.001} \\
GPT-2 Italian         & 1.5              & 0.359                     \\
UmBERTo               & 1.1              & 0.434                     \\
\textbf{It BERT Base} & \textbf{7.4}     & \textbf{0.019}            \\
\textbf{It BERT XXL}  & \textbf{9.1}     & \textbf{0.010}            \\
mBERT                 & 0.7              & 0.529                    \\
XLM-100               & \textless{}0.1              & 0.815                     \\
XLM-R Base            & \textless{}0.1   & 0.830                     \\
XLM-R Large           & 0.5              & 0.575                     \\
\textbf{XGLM 2.9B}    & \textbf{7.1}     & \textbf{0.022}           \\
\textbf{XGLM 4.5B}    & \textbf{12}     & \textbf{0.003}           \\
\textbf{XGLM 7.5B}    & \textbf{18.4}    & \textbf{\textless{}0.001} \\ \hline
\end{tabular}
\caption{Results of the ANOVAs for all models. Models for which there is a significant effect of the manipulation tested are shown in bold.}
\label{tab:exp4.2}
\end{table}

\newpage

\subsection*{Experiment 4: Argument x Person Interaction}

\begin{table}[!h]
\centering
\begin{tabular}{lrr}
\hline
\textbf{Model}     & \textbf{Chisq(df=1)} & \textbf{Corrected \textit{p}}     \\ \hline
GePeTto            & 2.7                  & 0.168          \\
GPT-2 Italian      & 0.2                  & 0.709           \\
UmBERTo            & 0.2                  & 0.739          \\
\textbf{It BERT Base }      & \textbf{5}                    & \textbf{0.048}          \\
It BERT XXL        & 0.3                  & 0.681          \\
mBERT              & 0.4                  & 0.607          \\
XLM-100            & \textless{}0.1       & 0.993          \\
XLM-R Base         & 1.1                  & 0.434          \\
XLM-R Large        & 0.8                  & 0.482           \\
\textbf{XGLM 2.9B}          & \textbf{8.9}                  & \textbf{0.008}         \\
\textbf{XGLM 4.5B} & \textbf{18.5}        & \textbf{\textless{}0.001} \\
\textbf{XGLM 7.5B} & \textbf{7.4}         & \textbf{0.015} \\ \hline
\end{tabular}
\caption{Results of the likelihood ratio tests for all models. Models for which there is a significant effect of the manipulation tested are shown in bold.}
\label{tab:exp4.1}
\end{table}

\end{document}